\documentclass[10pt,twocolumn,letterpaper]{article}

\pdfoutput=1

\usepackage{iccv}
\usepackage{times}
\usepackage{epsfig}
\usepackage{graphicx}
\usepackage{amsmath}
\usepackage{amssymb}

\usepackage{multirow}
\usepackage{makecell}
\usepackage{graphicx}
\usepackage{colortbl} 
\usepackage{arydshln} 
\usepackage{booktabs} 
\usepackage{array}
\usepackage{diagbox}
\usepackage{bbding}

\usepackage{overpic}
\usepackage{caption}
\usepackage{subcaption}
\usepackage{wrapfig}
\usepackage{graphicx}

\usepackage{xcolor}

\usepackage[pagebackref=true,breaklinks=true,letterpaper=true,colorlinks,bookmarks=false]{hyperref}

\iccvfinalcopy 

\ificcvfinal\fi

\begin{document}

\title{RPG-Palm: Realistic Pseudo-data Generation for Palmprint Recognition}

\author{Lei Shen\footnotemark[1] \\
Youtu Lab, Tencent\\
Shanghai, China\\
{\tt\small shenlei1996@gmail.com}
\and
Jianlong Jin\footnotemark[1]\\
Hefei University of Technology\\
Hefei, China\\
{\tt\small 2022111048@mail.hfut.edu.cn}
\and
Ruixin Zhang\\
Youtu Lab, Tencent\\
Shanghai, China\\
{\tt\small ruixinzhang@tencent.com}
\and
Huaen Li\\
Hefei University of Technology\\
Hefei, China\\
{\tt\small huaenli@mail.hfut.edu.cn}
\and
Kai Zhao\\
UCLA\\
California, USA\\
{\tt\small kz@kaizhao.net}
\and
Yingyi Zhang\\
Youtu Lab, Tencent\\
Shanghai, China\\
{\tt\small sherlyzhang@tencent.com}
\and
Jingyun Zhang\\
WeChat Pay Lab33, Tencent\\
Shenzhen, China\\
{\tt\small naskyzhang@tencent.com}
\and
Shouhong Ding\footnotemark[2]\\
Youtu Lab, Tencent\\
Shanghai, China\\
{\tt\small ericshding@tencent.com}
\and
Yang Zhao\footnotemark[2]\\
Hefei University of Technology\\
Hefei, China\\
{\tt\small YZhao@hfut.edu.cn}
\and
Wei Jia\footnotemark[2]\\
Hefei University of Technology\\
Hefei, China\\
{\tt\small jiawei@hfut.edu.cn}
}

\maketitle
\ificcvfinal\thispagestyle{empty}\fi
\renewcommand{\thefootnote}{\fnsymbol{footnote}} 
\footnotetext[1]{Equal contribution.} 
\footnotetext[2]{Corresponding authors.} 

\begin{abstract}
\vspace{-0.3cm}
Palmprint recently shows great potential in recognition applications as it is a privacy-friendly and stable biometric.
However, the lack of large-scale public palmprint datasets limits further research and development of palmprint recognition.
In this paper, we propose a novel realistic pseudo-palmprint generation (RPG) model to synthesize palmprints with massive identities. 
We first introduce a conditional modulation generator to improve the intra-class diversity. 
Then an identity-aware loss is proposed to ensure identity consistency against unpaired training. 
We further improve the B\'ezier palm creases generation strategy to guarantee identity independence. 
Extensive experimental results demonstrate that synthetic pretraining significantly boosts the recognition model performance. 
For example, our model improves the state-of-the-art B\'ezierPalm by more than $5\%$ and $14\%$ in terms of TAR@FAR=1e-6 under the $1:1$ and $1:3$ Open-set protocol. 
When accessing only $10\%$ of the real training data, our method still outperforms ArcFace with $100\%$ real training data, indicating that we are closer to real-data-free palmprint recognition.
\end{abstract}

\begin{figure}[!htb]
   \centering
   \center{\includegraphics[width=8cm]  {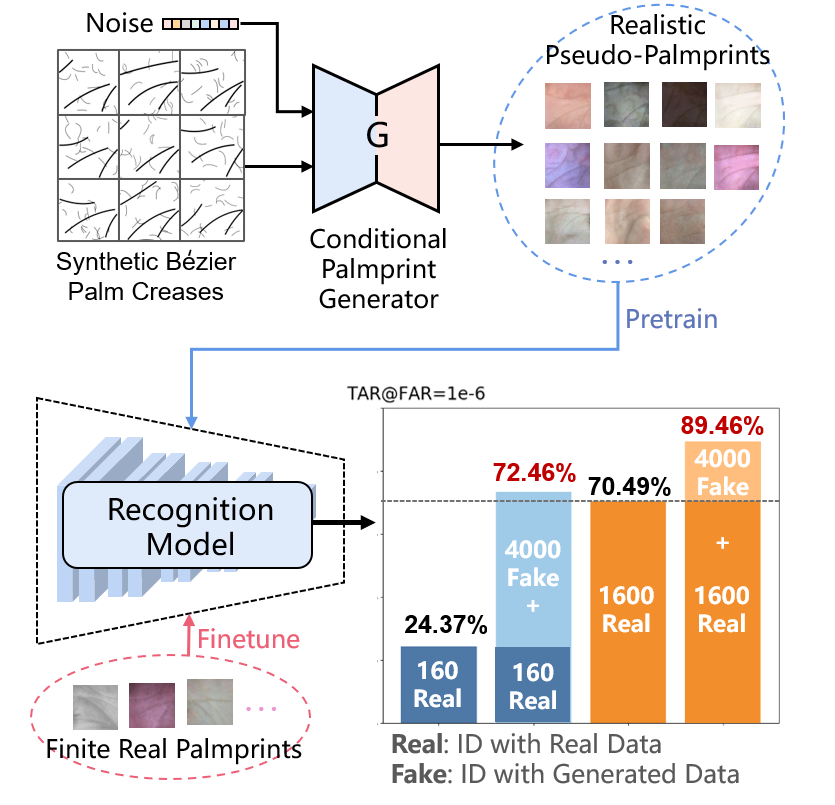}}
   \vspace{-0.2cm}
   \caption{
Synthetic B\'ezier palm creases are used as the identity control condition to guide the generator to produce diverse synthetic data. 
Recognition models are pretrained with pseudo-data and then finetuned on real data, achieving significant performance boosts. 
Besides, even if only accessing 10\% of the real training data for generation and recognition model, our method still gets comparable performance to 100\% of real data direct training.
   }
   \label{fig:into-fig}
\end{figure}

\section{Introduction}
Palmprint is an excellent biometric in terms of privacy, willingness, convenience, and security.
Since the palmprint locates on the inner side of the palm, obtaining the palmprint without one's permission is nearly impossible, 
which is more privacy-friendliness over the face. Compared to iris and fingerprints, palmprint has fewer usage restrictions, which makes palmprint more user-friendly. Furthermore, Palmprints and palm veins can be collected simultaneously to form a highly secure dual-modal system. 

Due to these advantages, big companies such as Amazon~\cite{amazonone} and Tencent~\cite{zhao2022bezier} begin to apply palmprint recognition in their payment services. 
However, for the same reason, palmprints are rare in public and expensive to collect. To our best knowledge, publicly available datasets~\cite{ceop, ferrer2011bispectral, hao2008multispectral, hassanat2015new, kanhangad2010contactless, kumar2008incorporating,shao2020effective, sun2005ordinal, zhang2020towards} only contain thousands of identities and tens of thousands of images in total. Meanwhile, face recognition has several million-level publicly available datasets~\cite{an2021partial, cao2018vggface2, guo2016ms, kemelmacher2016megaface, maze2018iarpa, whitelam2017iarpa} that contain tens to hundreds of thousands of identities. The lack of large-scale public dataset seriously inhibits the research on palmprint recognition.

To solve this problem, one way is to collect a large-scale palmprint dataset. 
However, this way is very time-consuming and expensive. 
In addition, collecting biometric data causes more and more privacy and ethical concerns~\cite{boutros2022sface} and is strictly restricted by legislation in many countries. 
Another way is to augment the training dataset with synthetic palmprints.
To better benefit palmprint recognition applications,  the synthetic method should support generating massive identities with inter-class and intra-class diversity,  and the gap between synthetic palmprints and real palmprints should be small.

Recently, B\'ezierPalm~\cite{zhao2022bezier} synthesizes fake palmprint images by generating palmprint creases with several parameterized B\'ezier curves for recognition model pretraining. 
B\'ezierPalm first shows the ability to output massive new identities without using real palmprints and significantly improves the performance of the palmprint recognition models. 
However, B\'ezierPalm has some problems unsolved.
Firstly, synthetic palmprints have a large gap to real ones, resulting in non-neglectable finetune data requirements in real-world applications. 
Secondly,  the curve parameter difference of each identity cannot ensure inter-class independence when generating massive data. 
Thirdly, the intra-class diversity only contains small curve deformation in B\'ezierPalm, while real palmprints have diverse textures, lighting and et al.

In this paper, we propose a realistic pseudo-palmprint generation (RPG) model. As shown in Fig.\ref{fig:into-fig}, the RPG model takes synthetic B\'ezier palm creases as an identity (ID) condition and outputs realistic pseudo-palmprints with a bidirectional mapping from the Gaussian noise domain to the palmprints domain.
Since there is no correspondence between synthetic B\'ezier palm creases and real palmprints, the generation model can only use unpaired data and will lose intra-class identity consistency easily during training. 
To solve this problem, we introduce an identity-aware loss that restrains the identity consistency between palmprints generated from the same B\'ezier palm creases. 
In addition, we design a conditional modulation generator to generate diversified intra-class textures and lighting conditions using a latent control vector encoded from random noises. 
To further reduce the distribution gap between synthetic palm creases and real-world palmprints, we refined the synthetic palmprint creases generation strategy with a more reasonable parameter design and identity independence check.
\vspace{0.3cm}

The contributions of this paper are as follows:\begin{itemize}
\item We propose a realistic pseudo-palmprint generation (RPG) model with a conditional modulation generator to improve the intra-class diversity and an ID-aware loss to help the RPG model ensure identity consistency under unpaired training. 
\item We improve the B\'ezier palm creases synthetic method to get more reasonable palm creases and independent identities.
\item Extensive experimental results on 13 public datasets demonstrate that recognition models pretrained with our synthetic pseudo-palmprints achieve state-of-the-art recognition accuracy. 
\item With our RPG pretraining, even if accessing only 10\% of the real data, the recognition model performance still outperforms 100\% real data direct training. Showing we are closer to real-data-free palmprint recognition.  
\end{itemize}

\section{Related Work}
\subsection{Palmprint Recognition Methods}
Palmprint recognition methods can be divided into two categories: traditional-based and deep-learning based~\cite{fei2018feature}. Traditional methods extract various kinds of local or global features to make different palmprints more discriminative. Local-based methods \cite{fei2016double,  guo2009palmprint, jia2008palmprint, kong2004competitive, luo2016local, sun2005ordinal, zheng2015suspecting} manually design effective local features for recognition. Global-based methods \cite{almeida2022ethics, feng2006alternative, hu2007two, lu2003palmprint, sang2009research, wang2006palmprint} extract low-dimensional features from the whole image to distinguish different palmprints. Deep Learning based methods ~\cite{dian2016contactless,  genovese2019palmnet, jia2022eepnet, svoboda2016palmprint} train modified neural networks to extract features with classification or pair-wise loss ~\cite{deng2019arcface, shen2022distribution, zhong2019centralized}. However, the lack of large-scale public palmprint datasets limits the potential of existing palmprint recognition methods. 

\subsection{Generation Model for Image Transformation}

With the development of image generation models, such as generative adversarial network (GAN)-based models~\cite{goodfellow2020generative, karras2019style} and diffusion-based models~\cite{ramesh2021zero, rombach2022high}, image-to-image generation/translation methods have achieved impressive performance~\cite{chen2020adversarial, choi2018stargan, isola2017image}. However, many typical models, e.g., conditional generation model pix2pixHD ~\cite{wang2018pix2pixHD}, multi-domain transformation model StarGAN~\cite{choi2018stargan}, dual-domain mapping model BicycleGAN~\cite{zhu2017toward} and recent conditional diffusion model~\cite{ saharia2022image, wang2022pretraining}, rely on paired training data, which is unavailable in many applications. 
Therefore, some unpaired image-to-image translation models have been proposed~\cite{kim2017learning, shao2021spatchgan, yi2017dualgan, zhu2017unpaired}, which usually adopt cycle consistency loss to train the models without paired data. However, these models are not designed for recognition tasks, and thus ignore the ID preservation in the generation process. Overall, there is still a lack of research on generating diversified and realistic palmprints with ID consistency from limited unpaired samples.

\subsection{Data generation for Recognition Tasks}

In order to improve the recognition performance, data generation can be used to expand the depth (diversity of each identity) and width (total number of identities) of training data. 
For example, in face recognition field, several facial image generation methods~\cite{deng2020disentangled, geng20193d, nguyen2019hologan, qiu2021synface, xi-towards-2017, fu2019dual, fu2021dvg} have been proposed to generate facial samples based on the priors of facial attributes, facial structures and 3D faces ~\cite{blanz1999morphable}. Also, in field of fingerprint recognition, some methods based on hand-crafted or learning-based approaches~\cite{bahmani2021high, engelsma2022printsgan, wyzykowski2021level} have been proposed to generate high-fidelity fingerprint images.

\begin{figure*}[htb]
   \centering
   \center{\includegraphics[width=0.95\linewidth]  {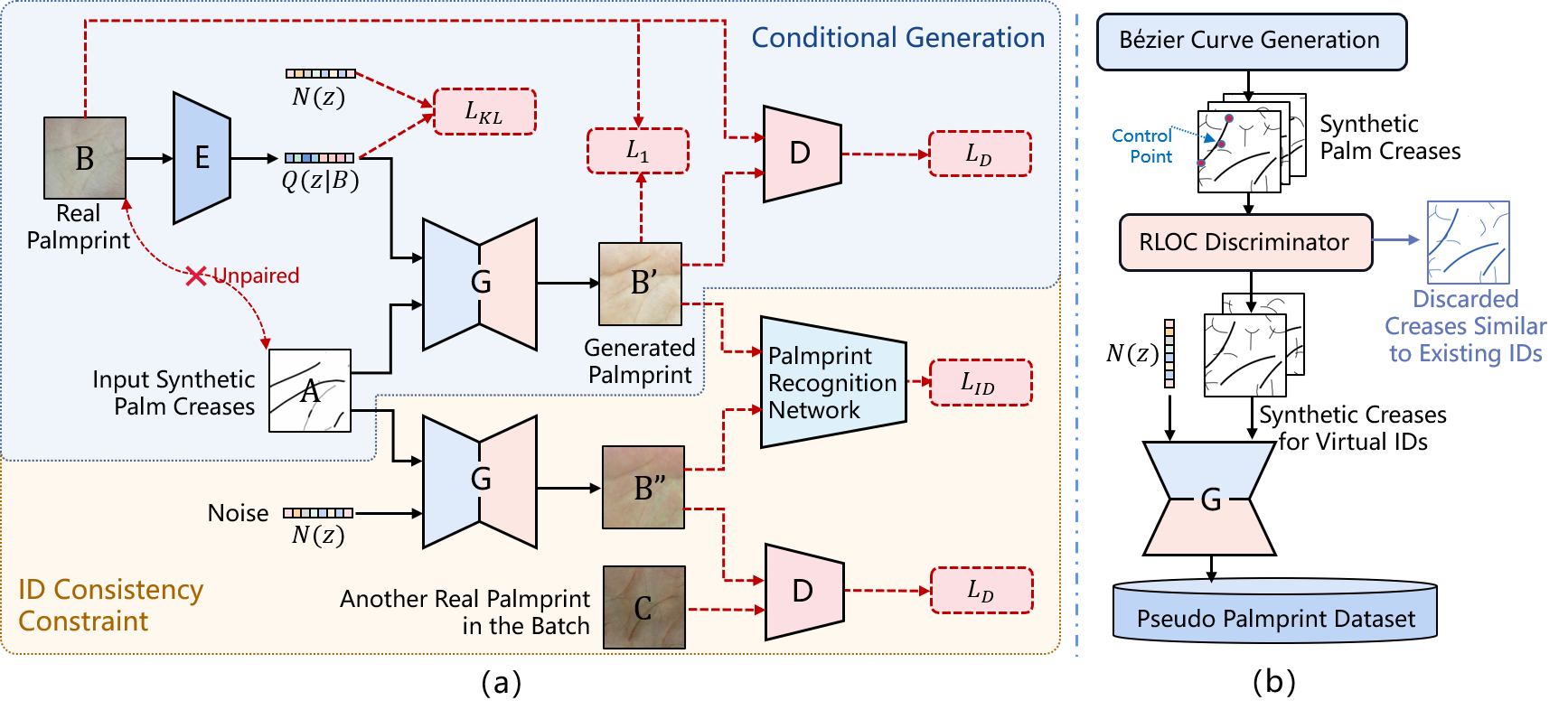}}
   \vspace{-0.6cm}
   \caption{
      Framework of the proposed realistic pseudo-palmprint generation model, (a) training stage, (b) forward pseudo-palmprint generation stage.
   }\label{fig:framework}
\end{figure*}

For palm recognition tasks, B\'ezierPalm~\cite{zhao2022bezier} use B\'ezier curves to synthesize fake palm creases by changing the B\'ezier control parameters. PalmGAN~\cite{shao2019palmgan} improves the cross-domain recognition performances by using CycleGAN to transfer the styles between normal and multispectral palmprints. However, PalmGAN doesn't create virtual identities, and the B\'ezierPalm suffers from the domain gap between palmprint images and geometry curves.
By introducing the new generation model and ID-aware loss, our method can expand the inter-class diversity and intra-class diversity of palmprint dataset simultaneously.
\vspace{-0.1cm}
\section{Method}
\subsection{Overall Framework}

Fig.\ref{fig:framework} illustrates the whole framework of the proposed realistic pseudo-palmprint generation (RPG) model, which includes a training stage and a forward palmprint generation stage. 
In the training phase, palmprint $B$ is first mapped from the palmprint image domain to the latent vector $Q(z|B)$ in the Gaussian noise domain through the encoder $E$.  
Then, the generator $G$ utilizes the unpaired palm creases $A$ as condition, and remaps the encoded noise vector $Q(z|B)$ back to the palmprint image domain. 
In order to increase the diversity and randomness of the generated palmprints, a conditional modulation structure is designed for $G$, which uses the input noise vector to control the modulation of intermediate features.

In order to constrain the ID consistency of the generated pseudo palmprints, an ID-aware loss is presented to enforce the generator to maintain the ID information of input palm creases.
As shown in Fig.\ref{fig:framework} (a), a siamese generator $G$ is used to produce another palmprint $B''$ with the same palm creases $A$. Then, a pretrained palmprint recognition discriminator is used to measure the ID consistency between $B'$ and $B''$. 

In the forward pseudo-palmprint generation stage, synthetic B\'ezier creases are input into the generator $G$ to obtain corresponding palmprint images for virtual identities.
Motivated by B\'ezierPalm ~\cite{zhao2022bezier}, we improve the random B\'ezier curves generation strategy to obtain a more reasonable layout of principal lines and wrinkle lines. Then, a classic palmprint recognition method RLOC ~\cite{jia2008palmprint} is applied to ensure the inter-class difference. 

In the following, we will introduce the details of generator $G$, encoder $E$, ID-aware loss, and the improved B\'ezier palm creases synthesis strategy.

\subsection{Conditional Modulation Palmprint Generator}

\begin{figure*}[!htb]
   \centering
   \center{\includegraphics[width=0.95\linewidth]  {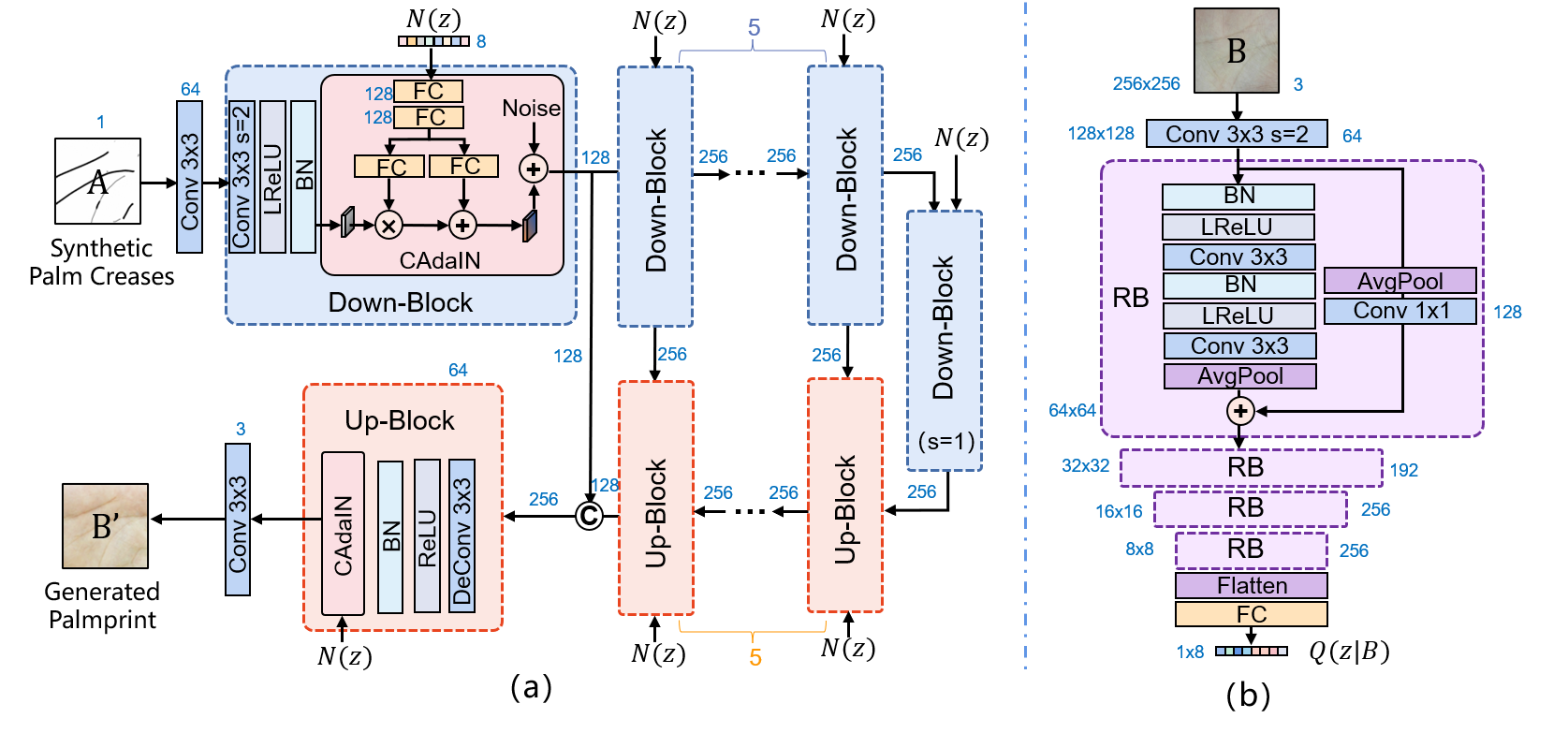}}
   \vspace{-0.4cm}
   \caption{
      The network structure of the proposed method, (a) structure of generator $G$, (b) structure of encoder $E$.
   }\label{fig:structure}
\end{figure*}

The proposed generator $G$ takes Gaussian noise vector as input, palm creases image $A$ as condition, and reproduces the pseudo-palmprint image $B'$. As a typical image generation task, the common UNet~\cite{ronneberger2015u} architecture is adopted. The detailed structure of $G$ is illustrated in Fig.\ref{fig:structure} (a). 
To generate diversified results, a conditional adaptive instance normalization module (CAdaIN) is introduced to modulate the generated details in each down-block and up-block. 

In the CAdaIN module, the input noise vector $N(z)$ is first encoded into a latent control vector $w(z)$ through two fully-connected (FC) layers. 
Note that the parameters of these two FC layers are shared for the whole generator $G$, so that the generated style can be consistent by means of the same $w(z)$.
Then, two other FC layers are used to modulate the mean and variance of intermediate feature maps respectively, which can be calculated as,
\begin{equation}
    \boldsymbol{X}_o = f_{c1}(w(z))\boldsymbol{X}_i + f_{c2}(w(z)) + \boldsymbol{n}_0 ,
\end{equation}
where $\boldsymbol{X}_i$ and $\boldsymbol{X}_o$ denote the input and output features of CAdaIN, and $f_{c1}$, $f_{c2}$ represent two FC layers. By using a latent control vector encoded from random noise, the CAdaIN can modulate the features to produce different distributions, resulting in diversified palmprint images. In addition, in order to further improve the diversity, random noise $\boldsymbol{n}_0$ with the same spatial resolution as the feature map is added to the modulated features to inject more randomness. 

For producing realistic results,  
the loss $\mathcal{L}_{1}$ and adversarial loss $\mathcal{L}_D$ are used to restrain the learning of $G$, as follows,
\begin{equation}
    \mathcal{L}_G=\lambda_1 \mathcal{L}_{1}(B, B')+\lambda_2 \mathcal{L}_D (B, B') ,
\end{equation}
where $\lambda_1$ and $\lambda_2$ are weights. The $\mathcal{L}_{1}$ is a commonly used pixel-wise loss, which can ensure the numerical similarity between the generated image and real palmprint. But note that the palm creases image $A$ is not matched with palmprint $B$, so that too strong pixel-wise constraint may cause wrong overfitting to the details in $B$. Hence, adversarial loss $\mathcal{L}_D$ is also used to relax the constraint and restrain the semantic similarity between $B$ and $B'$. For the discriminator in $\mathcal{L}_D$, we adopt the PatchGAN ~\cite{demir2018patch}, which determines patch-wise authenticity by mapping the image to a $70\times70$ grid.

\subsection{Palmprint Encoder}

The encoder $E$ is used to map the palmprint image to the Gaussian noise domain. Its structure is shown in Fig.\ref{fig:structure} (b), which is a straightforward ResNet structure. The original size of the feature map is $256\times 256$, and it gradually decreases to $16\times 16$ through several residual blocks (RB).
Then, a FC layer is used to estimate the target mean $\mu_Q$ and variance $\sigma^2_Q$ from flattened features. 
Finally, the noise vector $Q(z|B)$ is sampled from Gaussian space $\mathcal{N}(\mu_Q,\sigma^2_Q)$ via reparameterization trick~\cite{zhu2017toward}.

In the training stage, we constrain the KL divergence between $Q(z|B) \sim  \mathcal{N}(\mu_Q,\sigma^2_Q)$ and noise vector $N(z) \sim \mathcal{N}(0,1)$ sampled from Gaussian space, so that the target domain of $E$ can keep approximating to the standard normal distribution. This loss can be computed as,
\begin{equation}   
    \mathcal{L}_{KL}=-\frac{1}{2}(1 + \log \sigma_Q^2 - \sigma_Q^2 - \mu_Q^2) .
\end{equation}
In this paper, $E$ and $G$ are combined to learn a domain-to-domain mapping from unpaired training data, instead of using a fully supervised manner. 
Firstly, this structure can avoid the dependence on manual labeling of paired data.
Secondly, conditional bidirectional domain-to-domain mapping can reproduce random and realistic images and keep the information of palm creases condition as well.

\subsection{ID-aware Loss}
For training recognition models, the generated palmprints not only need to be diversified, but also have to preserve ID information. For a synthetic palm creases image $A$, the generated palmprints should have the same ID information. 
Therefore, an ID-aware loss $\mathcal{L}_{ID}$ is added to restrain the generator. As shown in Fig.\ref{fig:framework} (a), a siamese generator $G$ is used to produce another palmprint image $B''$ with the same creases $A$ and random noise vector $N(z)$. Then the $\mathcal{L}_{ID}$ restricts the ID consistency of two generated results of $B'$ and $B''$, as follows, 
\begin{equation}
\mathcal{L}_{ID}=1-\frac{D_{MB}(B')\cdot D_{MB}(B'')}{||D_{MB}(B')|| \times ||D_{MB}(B'')||} ,
\end{equation}
where ``$\cdot$'' denotes the vector dot product operation, $D_{MB}$ is a pretrained palmprint recognition model 
using MobileFaceNet~\cite{chen2018mobilefacenets} and 
extracts the $512\times 1$ feature of $B'$ and $B''$. 
That is, $\mathcal{L}_{ID}$ calculates cosine similarity between the extracted features of two generated images for the same ID.
Owing to the ID-aware loss, increasing the randomness and diversity of the generator will not destroy the intra-class ID consistency.

The total loss of entire model, as follows, 
\begin{equation}
\mathcal{L}_{total}= \lambda_D \mathcal{L}_{D}  + \lambda_1 \mathcal{L}_{1} + \lambda_{KL} \mathcal{L}_{KL} + \lambda_{ID} \mathcal{L}_{ID} ,
\end{equation}
where $\mathcal{L}_D$ denotes GAN loss, 
$\mathcal{L}_{1}$ denotes absolute error loss,
$\mathcal{L}_{KL}$ denotes KL divergence loss, 
and $\mathcal{L}_{ID}$ denotes ID-aware loss. 
We will add the total loss in final version.

\begin{figure}[tb]
   \centering
   \center{\includegraphics[width=1.0\linewidth]  {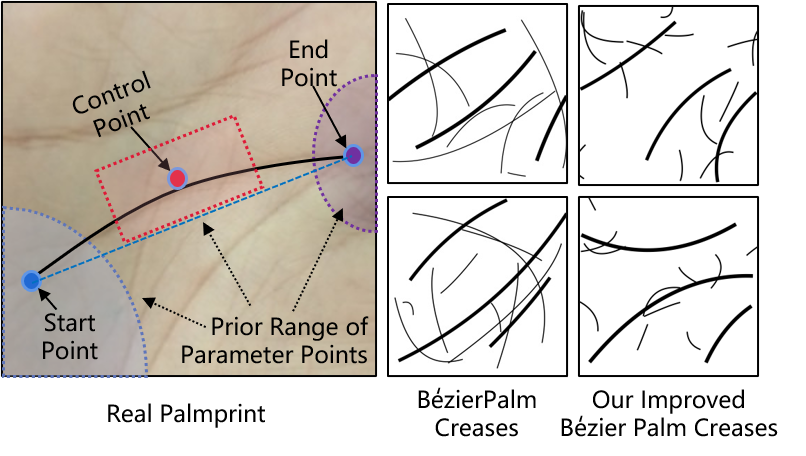}}
   \vspace{-0.9cm}
   \caption{
      Examples of some synthetic palm creases in B\'ezierPalm~\cite{zhao2022bezier} and our method.
   }\label{fig:creases}
\end{figure}

\subsection{Improved B\'ezier Palm Creases Synthesis}

In the forward stage, only the generator $G$ is used to produce palmprint images from synthetic palm creases, as shown in Fig.\ref{fig:framework} (b). Motivated by B\'ezierPalm~\cite{zhao2022bezier}, we also adopt the two-level B\'ezier curves to synthesize palm creases image that contains three principal lines and random wrinkle lines. The parametric control points of the B\'ezier curve can be adjusted randomly within a preset range to obtain a large number of fake palm creases for virtual ID. 

However, we have observed that randomly adjusted B\'ezier curves may lead to some unreasonable results. As shown in Fig.\ref{fig:creases}, the layout of some randomly generated creases is quite different from that of the real palmprints. 

As a result, this paper improves the B\'ezier curves synthesis strategy based on real palmprint prior from the following three aspects. 
Firstly, we observe and adjust the rough range of the parameter points (i.e., start point, control point, and end point) of three principal lines according to real palmprints. So that the layout of synthetic principal lines becomes more similar to that of real palms. 
Secondly, we adjust the synthesis rules to generate more wrinkle lines with moderate length and more uniform distribution. 
Thirdly, a similarity constraint based on RLOC~\cite{jia2008palmprint} is added, which makes the principal lines of different IDs sufficiently distinguishable. RLOC is a classic palmprint recognition method that can measure the similarity of two creases images. In order to avoid very similar principal lines being generated for different IDs, we filter out some creases images which exceed the RLOC inter-class similarity threshold. We ablate the threshold of RLOC with $0.1$, $0.5$, $0.9$, $0.95$ and experimentally set it as $0.9$. More details of improved B\'ezier palm creases can be found in supplementary materials.

Finally, the improved random B\'ezier curves are input into the generator $G$ to obtain corresponding palmprint images. This generation process can be repeated until a large-scale realistic pseudo-palmprint dataset is established. 
\section{Experimental Settings}
We adopt the same experimental datasets and Open-set evalutation protocol following B\'ezierPalm~\cite{zhao2022bezier}. The identities of the training and test set are isolated. TAR(True Acceptance Rate) @FAR(False Accept Rate) is used to evaluate model performance. Details about Open-set and evaluation protocol can be found in supplementary materials.
\subsection{Datasets}
We adopt $13$ public datasets in our experiments as in B\'ezierPalm~\cite{zhao2022bezier}, including $3,268$ IDs and $59,162$ palmprint images. Detailed descriptions of these datasets are shown in Tab.\ref{tab:dataset-stats}. We follow the detect-then-crop protocol in~\cite{zhang2019pay} to extract the Region Of Interests(ROIs). 
\begin{table}[!htbp]
   \centering
   \resizebox{0.9\linewidth}{!}{
   \renewcommand{\arraystretch}{1.1}
   \begin{tabular}{l|ccc}
    \hline
      Name & \#IDs & \#Images & Device \\
      \hline
      MPD~\cite{zhang2020towards}             & 400 & 16,000  & Phone \\
      TCD~\cite{zhang2020towards}             & 600 & 12,000  & Contactless \\
      IITD~\cite{kumar2008incorporating}      & 460 & 2,601 & Contactless \\
      CASIA~\cite{sun2005ordinal}             & 620 & 5,502 & Contactless\\
      CASIA-MS~\cite{hao2008multispectral}    & 200 & 7,200 & Contactless \\
      COEP~\cite{ceop}                        & 167 & 1,344 & Digital camera \\
      MOHI~\cite{hassanat2015new}             & 200 & 3,000 & Phone \\
      WEHI~\cite{hassanat2015new}             & 200 & 3,000 & Web cam \\
      GPDS~\cite{ferrer2011bispectral}        & 100 & 2,000 & Web cam \\          
      XJTU\_UP~\cite{shao2020effective}       & 200 & 30,000  & Phone \\
      XJTU\_A~\cite{shao2020effective}        & 114 & 1,130  & CMOS camera \\
      PolyU-MS~\cite{zhang2009online}         & 500 & 24,000 & Contactless \\
      PolyU(2D+3D)~\cite{kanhangad2010contactless} & 400 & 8,000 & Web cam \\ 
      \hline
   \end{tabular}}
   \vspace{-0.3cm}
   \caption{Details of the 13 public palmprint datasets.}
   \label{tab:dataset-stats}
\end{table}
\vspace{-0.3cm}

\subsection{Implementation Details} 
{\bf Generation Model Training.}
We generate 4000 identities and 100 samples for each identity by default following B\'ezierPalm~\cite{zhao2022bezier}. For training the generation model, the weights of $\mathcal{L}_{1}$, $\mathcal{L}_{D}$ and $\mathcal{L}_{KL}$ are set as $10.0$ $1.0$ and $0.01$ according to~\cite{zhu2017toward}. We ablate the weight of $\mathcal{L}_{ID}$ with $1.0$, $5.0$, $10.0$ and experimentally set it as $5.0$. The resolution of input and output images is $256\times 256$. The learning rate is 0.0002 in the first 30 epochs and linearly decays to 1e-8 in the last 30 epochs. The generation model is trained using Adam optimizer with parameters $(0.5, 0.99)$.  
For comparative experiment, we use the source codes of pix2pixHD~\cite{wang2018pix2pixHD}, CycleGAN~\cite{CycleGAN2017} and BicycleGAN~\cite{zhu2017toward} in their original papers. 
\begin{table*}[!htb]
   \centering
   \resizebox{0.9\linewidth}{!}{
   \begin{tabular}{lc|cccc|ccccc}
   \hline
   \multirow{3}{*}{Method} & \multirow{3}{*}{Backbone} & 
   \multicolumn{4}{c|}{train : test = 1 : 1} & \multicolumn{4}{c}{train : test = 1 : 3} \\
   &  & \makecell{TAR@ \\ 1e-3} & \makecell{TAR@ \\ 1e-4} & \makecell{TAR@ \\ 1e-5} & \makecell{TAR@ \\ 1e-6} &
                        \makecell{TAR@ \\ 1e-3} & \makecell{TAR@ \\ 1e-4} & \makecell{TAR@ \\ 1e-5} & \makecell{TAR@ \\ 1e-6} \\
   \hline
   CompCode~\cite{kong2004competitive}     & N/A & 0.4800 & 0.4292 & 0.3625 & 0.2103 & 0.4501 & 0.3932 & 0.3494 & 0.2648 \\
   LLDP~\cite{luo2016local}                & N/A & 0.7382 & 0.6762 & 0.5222 & 0.1247 & 0.7372 & 0.6785 & 0.6171 & 0.2108 \\
   BOCV~\cite{guo2009palmprint}              & N/A & 0.4930 & 0.4515 & 0.3956 & 0.2103 & 0.4527 & 0.3975 & 0.3527 & 0.2422 \\
   RLOC~\cite{jia2008palmprint}            & N/A & 0.6490 & 0.5884 & 0.4475 & 0.1443 & 0.6482 & 0.5840 & 0.5224 & 0.3366 \\
   DOC~\cite{fei2016double}                & N/A & 0.4975 & 0.4409 & 0.3712 & 0.1667 & 0.4886 & 0.4329 & 0.3889 & 0.2007 \\
   PalmNet~\cite{genovese2019palmnet}      & N/A & 0.7174 & 0.6661 & 0.5992 & 0.1069 & 0.7217 & 0.6699 & 0.6155 & 0.2877 \\
   \hline
   C-LMCL~\cite{zhong2019centralized}      & MB  & 0.9290 & 0.8554 & 0.7732 & 0.6239 & 0.8509 & 0.7554 & 0.7435 & 0.5932 \\
   ArcFace~\cite{deng2019arcface}          & MB  & 0.9292 & 0.8568 & 0.7812 & 0.7049 & 0.8516 & 0.7531 & 0.6608 & 0.5825 \\
   B\'ezierPalm~\cite{zhao2022bezier}      & MB  & 0.9640 & 0.9438 & 0.9102 & 0.8437 & 0.9407 & 0.8861 & 0.7934 & 0.7012 \\
   Ours                                    & MB  & \textbf{0.9802} & \textbf{0.9714} & \textbf{0.9486} & \textbf{0.8946} & 
                                                   \textbf{0.9496} & \textbf{0.9267} & \textbf{0.8969} & \textbf{0.8485} \\
   \hline
   C-LMCL~\cite{zhong2019centralized}      & R50 & 0.9545 & 0.9027 & 0.8317 & 0.7534 & 0.8601 & 0.7701 & 0.6821 & 0.6254 \\
   ArcFace~\cite{deng2019arcface}          & R50 & 0.9467 & 0.8925 & 0.8252 & 0.7462 & 0.8709 & 0.7884 & 0.7156 & 0.6580 \\
   B\'ezierPalm~\cite{zhao2022bezier}      & R50 & 0.9671 & 0.9521 & 0.9274 & 0.8956 & 0.9424 & 0.8950 & 0.8217 & 0.7649 \\
   Ours                                    & R50 & \textbf{0.9821} & \textbf{0.9732} & \textbf{0.9569} & \textbf{0.9347} & 
                                                   \textbf{0.9533} & \textbf{0.9319} & \textbf{0.9016} & \textbf{0.8698} \\
   \hline
   \end{tabular}
   }
   \vspace{-0.3cm}
   \caption{
      Quantitative performances under the open-set protocol where the performances
      are evaluated in terms of TAR@FAR.
      `MB' represents the MobileFaceNet~\cite{chen2018mobilefacenets} backbone
      and `R50' is resnet50~\cite{he2016deep} backbone.}
   \label{tab:openset}
\end{table*}

{\bf Recognition Model Training.}
For our recognition model, we use ResNet50~\cite{he2016deep} and MobileFaceNet~\cite{chen2018mobilefacenets} as the backbone with the input resolution of $224 \times 224$. The model is first pretrained on synthesized data for 25 epochs and then finetuned on real datasets for 50 epochs. For the baseline model, we train the model on real datasets for 50 epochs. The feature extractor in ID-aware Loss uses the same training setting as the baseline. We use Arcface~\cite{deng2019arcface} with margin $m=0.5$ and scale factor $s=48$ for the pretraining, finetuning and baseline training supervision. We use the cosine learning rate schedule with a warmup start for one epoch. The maximum learning rate is 1e-2 and the minimum learning rate is 1e-6 for pretraining and finetuning. All recognition models are trained with a mini-batch SGD optimizer. We use 4 NVIDIA Tesla V100 GPUs for training with total batch size of $128$.

It should be emphasized that the generation model, the feature extractor in ID-aware loss, and the recognition model all ensure that the training and test sets are completely isolated.
\section{Experimental Results}
\subsection{Open-set Palmprint Recognition}
\begin{figure}[!htb]
   \centering
   \begin{overpic}[width=1.0\linewidth]{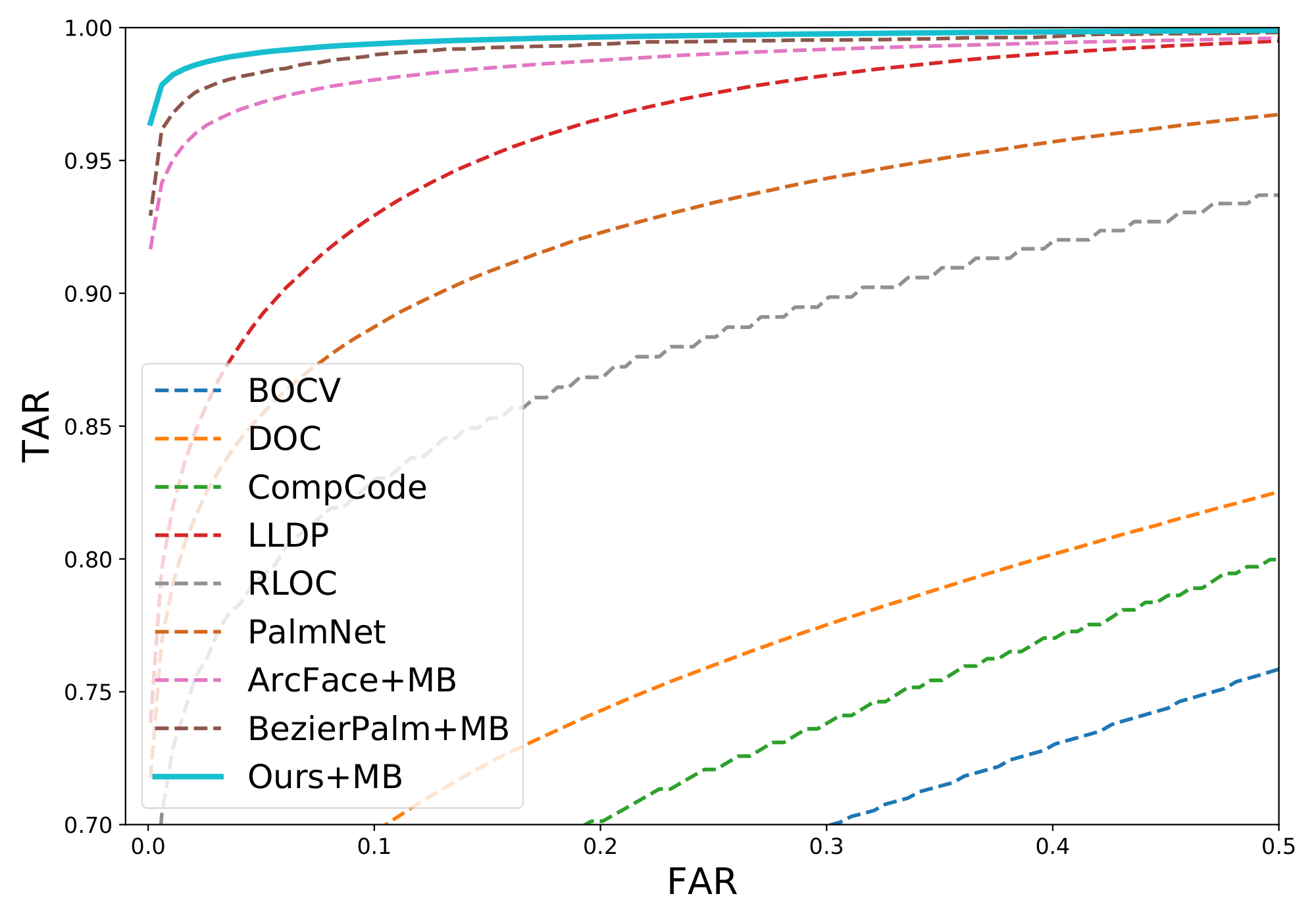}
   \end{overpic}
   \vspace{-0.9cm}
   \caption{
      TAR@FAR curves of different methods under the open-set 1:1 setting. The backbone of ArcFace, B\'ezierPalm and our method is MobileFaceNet. 
   }\label{fig:openset-tar-far}
\end{figure}

We first test our method under the open-set protocol with two different training and test ratios 1:1 and 1:3 (trainIDs:testIDs=1634:1632, 818:2448). Details about the "Open-set" protocol can be found in supplementary materials.
The quantitative results are shown in Tab.\ref{tab:openset}. The TAR $v.s.$ FAR curves of the 1:1 setting are in Fig.\ref{fig:openset-tar-far}. Our method outperforms B\'ezierPalm by $5.09\%$ and $14.73\%$ under 1:1 and 1:3 settings @FAR=1e-6 using `MB', which establishes a new state-of-the-art. Our method achieves more significant improvement under 1:3 setting than 1:1 setting, which demonstrates the effectiveness of our method in scenarios with only a small amount of real training data.
\subsection{Palmprint Recognition with Limited Identities}
To further verify the performance of our method with limited real training IDs, we test our method with different sizes of real training IDs and fix the test set under the 1:1 open-set protocol. We synthesize 4000 identities and 100 samples for each identity in all experiments.
As shown in Tab.\ref{tab:limited-identities}, the performance of ArcFace and B\'ezierPalm drops quickly with the reduction of real data, while our method still performs well with few real training IDs. Note that our method with $160(10\%)$ real IDs even outperforms Arcface with $1600$ real IDs. Besides, our method significantly improves the TAR by 20.81\%@FAR=1e-6 against B\'ezierPalm~\cite{zhao2022bezier} with $80$ training IDs, which shows the superiority of our method under few real training IDs.
\begin{table}[!htb]
   \centering
   \setlength\tabcolsep{1.5mm}
   \renewcommand{\arraystretch}{1.2}
   \resizebox{0.9\linewidth}{!}{
        \begin{tabular}{lc|cccc}
         \hline
        \multirow{2}{*}{Method} & \multirow{2}{*}{\#ID} & \multicolumn{4}{c}{TAR@FAR=} \\
        & & 1e-3 & 1e-4 & 1e-5 & 1e-6 \\
        \hline
        ArcFace & \multirow{3}{*}{1,600} & 0.9292 & 0.8568 & 0.7812 & 0.7049 \\
        B\'ezierPalm   &  & 0.9640 & 0.9438 & 0.9102 & 0.8437 \\
        Ours           &  &                \textbf{0.9802} & \textbf{0.9714} & \textbf{0.9486} & \textbf{0.8946} \\
        \hline    
        ArcFace & \multirow{3}{*}{800} &   0.8934 & 0.7432 & 0.7104 & 0.6437 \\
        B\'ezierPalm   &  & 0.9534 & 0.9390 & 0.9025 & 0.8164 \\
        Ours           &  &                \textbf{0.9783} & \textbf{0.9687} & \textbf{0.9356} & \textbf{0.8741} \\
        \hline
        ArcFace  & \multirow{3}{*}{400} &  0.8102 & 0.7050 & 0.6668 & 0.3320 \\
        B\'ezierPalm   &  & 0.9189 & 0.8497 & 0.7542 & 0.6899 \\
        Ours           &  &                \textbf{0.9573} & \textbf{0.9324} & \textbf{0.8836} & \textbf{0.8162} \\
         \hline
        ArcFace  & \multirow{3}{*}{160} & 0.6761 & 0.5294 & 0.4783 & 0.2437 \\
        B\'ezierPalm   &  & 0.8179 & 0.6998 & 0.5826 & 0.4832 \\
        Ours           &  &                \textbf{0.9356} & \textbf{0.8641} & \textbf{0.8063} & \textbf{0.7246} \\
        \hline
        ArcFace  & \multirow{3}{*}{80} & 0.5384 & 0.4682 & 0.3249 & 0.1173 \\
        B\'ezierPalm   &   & 0.6547 & 0.5511 & 0.4490 & 0.3743 \\
        Ours           &  &                \textbf{0.8974} & \textbf{0.8092} & \textbf{0.6947} & \textbf{0.5824} \\
         
        \hline
        \end{tabular}
   }
   \vspace{-0.3cm}
   \caption{
      Performance under different real training identities.
      The generation model, feature extractor in ID-aware loss, and the recognition model access the same number of real training identities.
      The backbone is MobileFaceNet. 
   }
   \label{tab:limited-identities}
\end{table}
\subsection{Palmprint Recognition at Million Scale}
In order to verify the effectiveness of our method on large-scale real-world datasets, we test our method on our internal dataset with millions of palmprint images. Our dataset contains $19,286$ training identities with $2,871,073$ images and $1,000$ test identities with $182,732$ images. For a fair comparison with B\'ezierPalm, we generate $20,000$ IDs with $100$ samples in each ID for pretraining. Quantitative results are shown in Tab.\ref{tab:million}. Our method outperforms ArcFace and B\'ezierPalm with a clear margin, showing its practical application value on large-scale datasets.
%
%
\begin{table}
    \centering
    \setlength\tabcolsep{1mm}
    \renewcommand{\arraystretch}{1.1}
    \resizebox{0.9\linewidth}{!}{
    \begin{tabular}{lc|cccccc}
     \hline
    \multirow{2}{*}{Method} & \multirow{2}{*}{Backbone} & \multicolumn{6}{c}{TAR@FAR=} \\
    & & 1e-6 & 1e-7 & 1e-8 & 1e-9 \\
    \hline
    ArcFace        & \multirow{3}{*}{MB}    & 0.9770 & 0.9550 & 0.9251 & 0.8833 \\
    B\'ezierPalm   &                        & 0.9803 & 0.9605 & 0.9301 & 0.9015 \\
    Ours           &                        & \textbf{0.9871} & \textbf{0.9684} & \textbf{0.9416} & \textbf{0.9194} \\
    \hline
    ArcFace        & \multirow{3}{*}{R50}   & 0.9986 & 0.9964 & 0.9931 & 0.9879 \\
    B\'ezierPalm   &                        & 0.9996 & 0.9975 & 0.9943 & 0.9911 \\
    Ours           &                        & \textbf{0.9998} & \textbf{0.9983} & \textbf{0.9972} & \textbf{0.9954} \\
    \hline
    \end{tabular}
    }
    \vspace{-0.3cm}
    \caption{
       Palmprint recognition performance on million scale dataset.
    }
    \label{tab:million}
 \end{table}

\section{Ablation Study}
In this section, we conduct ablation studies to verify different components and design choices of our method. The MobileFaceNet is used as the backbone for all experiments under the same Open-set protocol.
\begin{figure*}[!htb]
   \centering
   \begin{overpic}[width=0.95\linewidth]{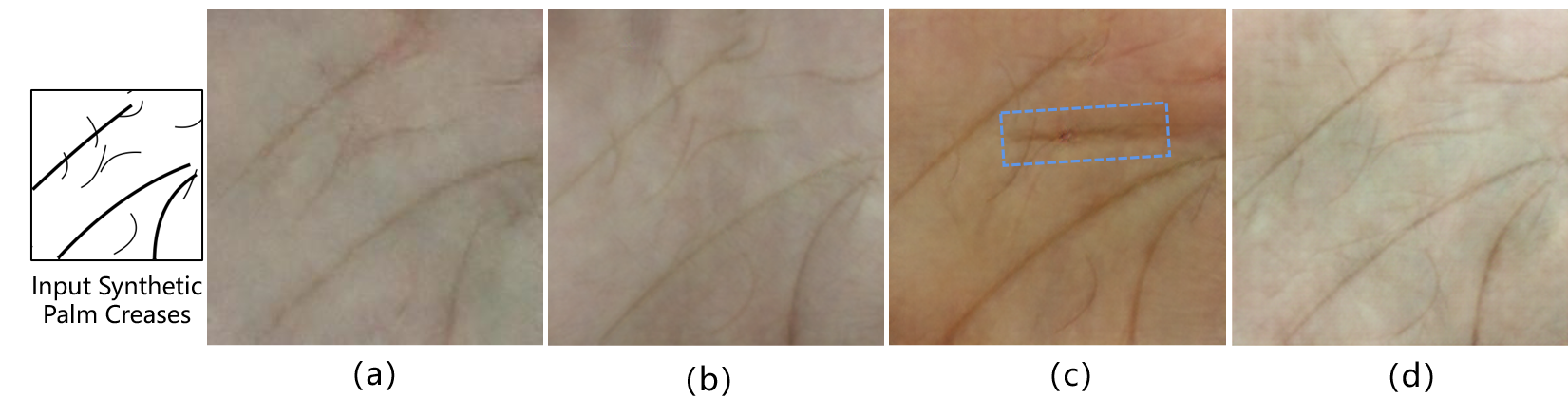}
   \end{overpic}
   \vspace{-0.6cm}
   \caption{
      Generated palmprint images of different methods, (a) pix2pixHD~\cite{wang2018pix2pixHD}, (b) CycleGAN~\cite{CycleGAN2017}, (c) BicycleGAN~\cite{zhu2017toward}, (d) the proposed method.
   }\label{fig:generated-palms}
\end{figure*}

\begin{figure}[!htb]
   \centering
   \center{\includegraphics[width=8cm]  {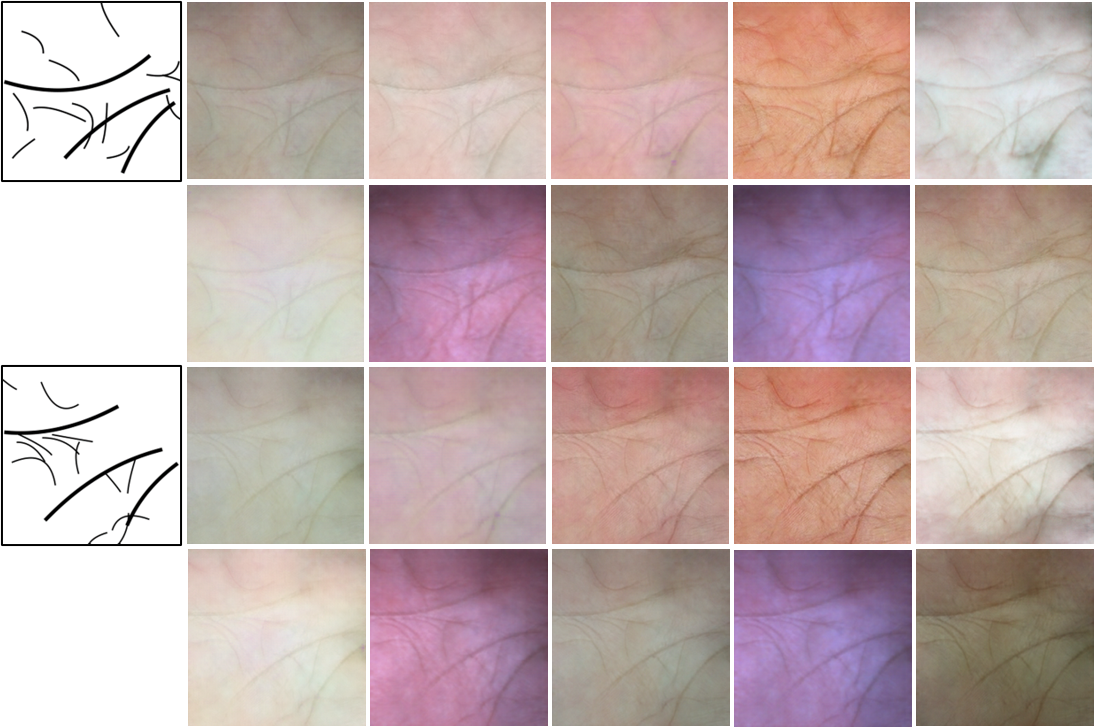}}
   \vspace{-0.2cm}
   \caption{
    Diverse synthetic palmprints can be generated by adjusting the input noise vector $N(z)$.
   }\label{fig:nz-fig}
\end{figure}
\subsection{Components and design choices}
The main components and design choices of our method are ID-aware loss, conditional modulation generator and improved strategy for B\'ezier palm creases synthesis. Tab.\ref{tab:components} shows the results of recognition models with or without these components. `I', `C', `S' represent the three components respectively. 
For baseline without `C', the normal UNet~\cite{ronneberger2015u} as in \cite{zhu2017toward} is used to take place of `C'.
ID-aware loss achieves the greatest improvement by 11.72\%@FAR=1e-6 against the baseline, which reflects its advantage for preserving intra-class consistency. 
The conditional modulation generator also significantly improves the performance by 4.49\%@FAR=1e-6, which reflects its advantage for improving the intra-class diversity. The improved B\'ezier palm creases bring an improvement by 1.15\%@FAR=1e-6.
\begin{table}[!htb]
   \centering
   \resizebox{0.9\linewidth}{!}{
        \begin{tabular}{ccc|cccc}
         \hline
            \multirow{2}{*}{I} & \multirow{2}{*}{C} & \multirow{2}{*}{S} & \multicolumn{4}{c}{TAR@FAR=} \\
            & & & 1e-3 & 1e-4 & 1e-5 & 1e-6 \\
            \hline
            \XSolidBrush & (UNet) & \XSolidBrush & 0.9399 & 0.8905 & 0.8164 & 0.7210 \\
            \CheckmarkBold     &  (UNet) & \XSolidBrush    & 0.9687 & 0.9571 & 0.9076 & 0.8382 \\
            \CheckmarkBold &\CheckmarkBold & \XSolidBrush    & 0.9796 & 0.9689 & 0.9441 & 0.8831 \\
            \CheckmarkBold &\CheckmarkBold & \CheckmarkBold  & \textbf{0.9802} & \textbf{0.9714} & \textbf{0.9486} & \textbf{0.8946} \\
            \hline
        \end{tabular}
        }
        \vspace{-0.3cm}
        \caption{
        Ablation of different components in our method. `I', `C' and `S' denote ID-aware loss, conditional modulation generator and improved synthetic creases, respectively.
   }
   \label{tab:components}
\end{table}
%
\begin{table}[!htb]
   \centering
   \renewcommand{\arraystretch}{1.3}
   \resizebox{1.0\linewidth}{!}{
   \begin{tabular}{l|c|cc|cc}
    \hline
   \multirow{3}{*}{Method} & \multirow{3}{*}{FID$\downarrow$} &
   \multicolumn{2}{c|}{train:test=1:1} & \multicolumn{2}{c}{train:test=1:3} \\
   & &  \makecell{TAR$\uparrow$ \\ @1e-5} & \makecell{TAR$\uparrow$ \\ @1e-6} & \makecell{TAR$\uparrow$ \\ @1e-5} & \makecell{TAR$\uparrow$ \\ @1e-6} \\
   \hline
   ArcFace~\cite{deng2019arcface}            & - - & 0.7812 & 0.7049 & 0.6608 & 0.5825 \\
   B\'ezierPalm~\cite{zhao2022bezier}        & - - & 0.9102 & 0.8437 & 0.7934 & 0.7012 \\
   pix2pixHD~\cite{wang2018pix2pixHD}        & 97.5801 & 0.9156 & 0.8734 & 0.8052 & 0.7244 \\
   CycleGAN~\cite{CycleGAN2017}              & 50.7704 & 0.9136 & 0.8703 & 0.7863 & 0.7189 \\
   BicycleGAN~\cite{zhu2017toward}           & 35.2801 & 0.8173 & 0.7254 & 0.6783 & 0.5929 \\
   Ours                                      & \textbf{16.4762} & \textbf{0.9486} & \textbf{0.8946} & 
                                                  \textbf{0.8969} & \textbf{0.8485} \\
   \hline
   \end{tabular}
   }
   \vspace{-0.3cm}
   \caption{
      Quantitative recognition results using different generation methods under the open-set protocol.}
   \label{tab:alation-generation-methods}
\end{table}
\subsection{Comparison of Generation Methods}
Three generation methods pix2pixHD~\cite{wang2018pix2pixHD}, CycleGAN~\cite{CycleGAN2017}, BicycleGAN~\cite{zhu2017toward} are used for comparison, and all of them are retrained on the same training set with unpaired data. 
Fig.\ref{fig:generated-palms} illustrates the generated palmprints of different generation methods. It can be found that pix2pixHD and CycleGAN tend to synthesize blurred results. As marked in blue rectangle, BicycleGAN may generate some creases that are inconsistent with the input B\'ezier curves. 
Our method can synthesize clearer and sharper principal lines and faithfully preserve the ID information of the input B\'ezier curves. 
For intra-class diversity, as shown in Fig.\ref{fig:nz-fig}, our method is able to randomly generate high-fidelity palmprints with diverse lighting and skin types by adjusting the input noise vector $N(z)$. 

Quantitative results are shown in Tab.\ref{tab:alation-generation-methods}. Our method substantially outperforms the existing methods in terms of TAR@FAR. Also, we use FID~\cite{bynagari2019gans} to evaluate the quality of generated results and our method also effectively decreases the FID score by more than $50\%$ against other methods. Besides, we reproduced some other generation methods~\cite{shao2021spatchgan, su2022dual} for unpaired image-to-image transfer, including diffusion-based models, but the generated results are unsatisfying. Related details and more subjective results can be found in supplementary material.
\subsection{Number of Synthesized Identities and Images}
\begin{figure}[!htbp]
   \centering
   \begin{overpic}[width=1.0\linewidth]{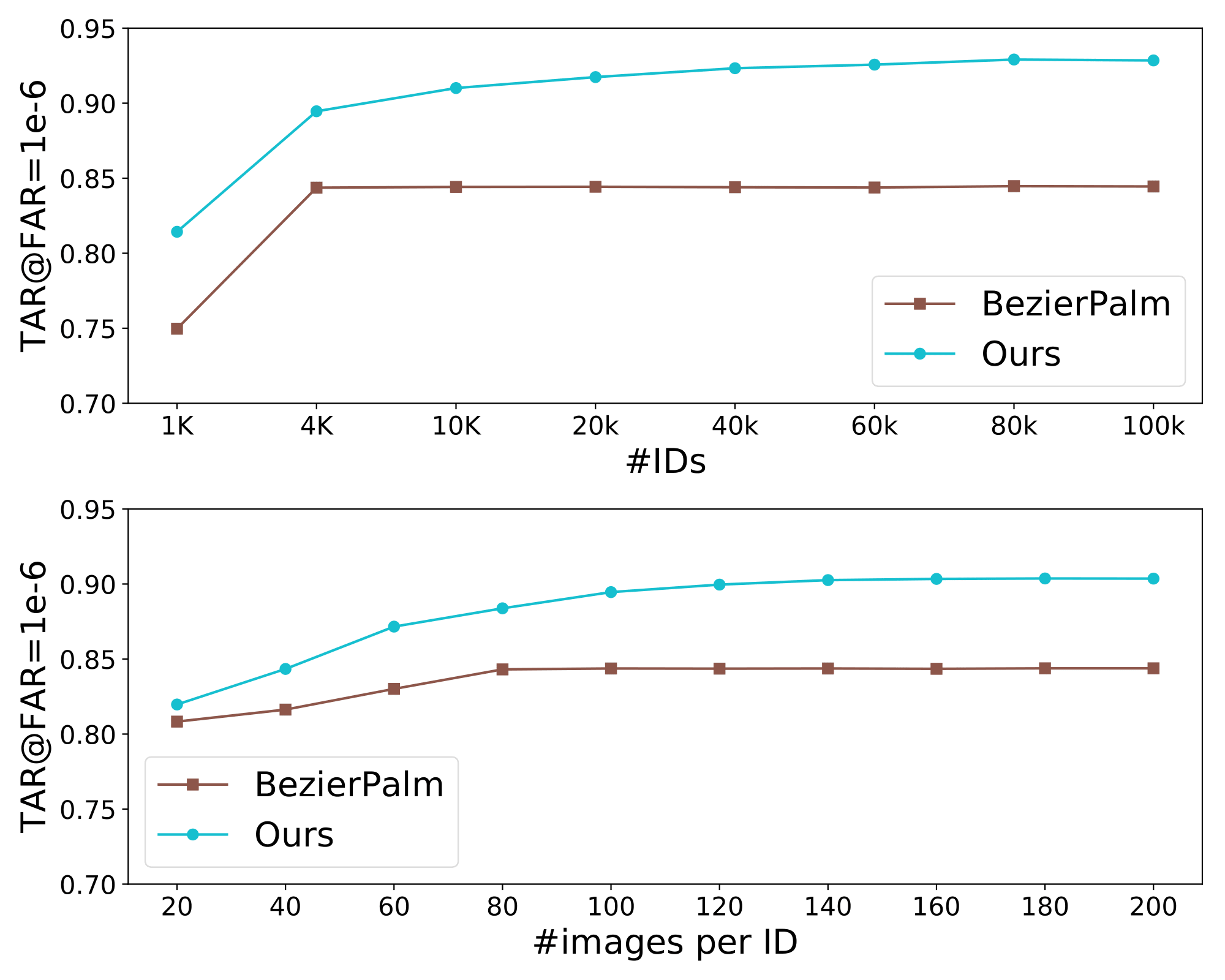}
   \end{overpic}
   \vspace{-0.9cm}
   \caption{
      TAR@FAR=1e-6 of recognition models pretrained with different numbers of synthetic identities and samples. The backbone is MobileFaceNet.
   }\label{fig:ablate-ids-identities}
\end{figure}

In this ablation, we investigate the influence of the number of synthesized identities and images. Specifically, we generate 4000 identities and 100 samples by default, and fix one number and vary the other under the open-set protocol(train:test=1:1).
The results are shown in Fig.\ref{fig:ablate-ids-identities}. With the increase of ``width" and ``depth" of synthetic data, our method can continuously improve the performance of the recognition model and achieve a higher upper bound than B\'ezierPalm. The performance reaches the upper bound with $80k$ synthetic identities and $160$ samples per identity. 
%
\section{Conclusion}
This paper proposed an ID-aware conditional modulation generation model which can produce realistic and diversified palmprint images. Specifically, a conditional modulation generator was designed, which adopted synthetic palm creases as condition, and used encoded Gaussian noise vector to modulate the generated diversity. An ID-aware loss was proposed to preserve the identity information of input palm creases during the unpaired training process. In the forward pseudo-palmprint generation stage, we improved the B\'ezier curves generation strategy to produce more realistic synthetic palm creases. 
From experimental results, we can obtain the following findings. Firstly, the generated pseudo-palmprint samples can effectively improve the performance of palmprint recognition models. Secondly, by using the synthetic palmprints, our method can effectively reduce the dependence of real data by $90\%$.
In future work, we hope to implement complete real-data-free palmprint recognition with synthetic data.

\section{Acknowledgements}
This work is partly supported by the grants of the National Natural Science Foundation of China, Nos.62076086, 62272142 and 61972129.

{\small
\bibliographystyle{ieee_fullname}
\bibliography{egpaper_for_review}
}

\end{document}